\documentclass[10pt,twocolumn,letterpaper]{article}
\usepackage{iccv}
\usepackage[square,sort,comma,numbers]{natbib}

\newcommand{\x}{\mathbf{x}}
\usepackage{booktabs}

\usepackage{times} 
\usepackage{epsfig} 
\usepackage{graphicx} 
\usepackage{amsmath} 
\usepackage{amssymb} 
\usepackage{algorithm}
\usepackage{algorithmic}

\usepackage{mdwtab}
\usepackage{tabularx,colortbl}
\usepackage{xspace}

\makeatletter

\DeclareRobustCommand\onedot{\futurelet\@let@token\@onedot}
\def\@onedot{\ifx\@let@token.\else.\null\fi\xspace}
 
\def\ie{\emph{i.e}\onedot} 
 
\def\etc{\emph{etc}\onedot} 
 
\def\etal{\emph{et al}\onedot}
\makeatother

\usepackage[pagebackref=true,breaklinks=true,letterpaper=true,colorlinks,bookmarks=false]{hyperref}
\iccvfinalcopy 

\begin{document} 

\title{Generic Object Detection with Dense Neural Patterns and Regionlets}


\author{
Will Y. Zou$^{\dagger\ddagger}$\,\,\,\,\,\,\,\,\,\,\,
Xiaoyu Wang$^\dagger $\,\,\,\,\,\,\,\,\,\,\,
Miao Sun$^{\dagger *} $\,\,\,\,\,\,\,\,\,\,\,
Yuanqing Lin$^\dagger $\,\,\,\,\,\,\,\,\,\,\,\\
  $^\dagger $NEC Laboratories America, Inc. \,\,\,\,\,\,\,\,$^{\ddagger}$Stanford University\,\,\,\,\,\,\,\, $^*$University of
 Missouri\\
 {\tt\small wzou@stanford.edu\,\,\,\,\,\,\,\,
fanghuaxue@gmail.com\,\,\,\,\,\,\,\,
  msqz6@mail.missouri.edu \,\,\,\,\,\,\,\,
ylin@nec-labs.com\,\,\,\,\,\,\,\,}\\
}

\maketitle
\begin{abstract} 

This paper addresses the challenge of establishing a bridge between deep convolutional neural
networks and conventional object detection frameworks for accurate and efficient generic object
detection.  We introduce Dense Neural Patterns, short for DNPs, which are dense local
features derived from discriminatively trained deep convolutional neural networks. DNPs can be
easily plugged into conventional detection frameworks in the same way as other dense local
features(like HOG or LBP).  The effectiveness of the proposed approach is demonstrated with
the Regionlets object detection framework.  It achieved 46.1\% mean average precision on the PASCAL VOC
2007 dataset, and 44.1\% on the PASCAL VOC 2010 dataset, which dramatically improves the original
Regionlets approach without DNPs.

\end{abstract} 

\section{Introduction} 
Detecting generic objects in high-resolution images is one of the most valuable pattern recognition
tasks, useful for large-scale image labeling, scene understanding, action recognition, self-driving
vehicles and robotics. At the same time, accurate detection is a highly challenging task due to
cluttered backgrounds, occlusions, and perspective changes. Predominant approaches~\cite{lsvm-pami}
use deformable template matching with hand-designed features.  However, these methods are not
flexible when dealing with variable aspect ratios. Wang \etal recently proposed a radically
different approach, named \emph{Regionlets}, for generic object detection~\cite{wang2013}. It extends
classic cascaded boosting classifiers ~\cite{Viola01} with a two-layer feature extraction hierarchy
which is dedicatedly designed for region based object detection.  The innovative framework is
capable of dealing with variable aspect ratios, flexible feature sets, and improves upon Deformable
Part-based Model by 8\%~\cite{wang2013} in terms of mean average precision. Despite the success of these sophisticated detection
methods, the features employed in these frameworks are still traditional features based on low-level
cues such as histogram of oriented gradients(HOG)~\cite{dalal2005histograms}, local binary
patterns(LBP)~\cite{AphonenPAMI06} or covariance~\cite{Tuzel:2008} built on image gradients. 


As with the success in large scale image classification~\cite{Krizhevsky12}, object detection using
a deep convolutional neural network also shows promising performance~\cite{Ross2013,
SermanetEZMFL13}.  The dramatic improvements from the application of deep neural networks are
believed to be attributable to their capability to learn hierarchically more complex features from
large data-sets.  Despite their excellent performance, the application of deep CNNs has been
centered around image classification, which is computationally expensive when transferring to object
detection. For example, the approach in~\cite{Ross2013} needs around 2 minutes to evaluate one
image. Furthermore, their formulation of the problem does not take advantage of venerable and
successful object detection frameworks such as DPM or \emph{Regionlets}  which are powerful designs for
modeling object deformation, sub-categories and multiple aspect ratios.

\begin{figure}[tb] 
\centering 
\includegraphics[width=1\columnwidth]{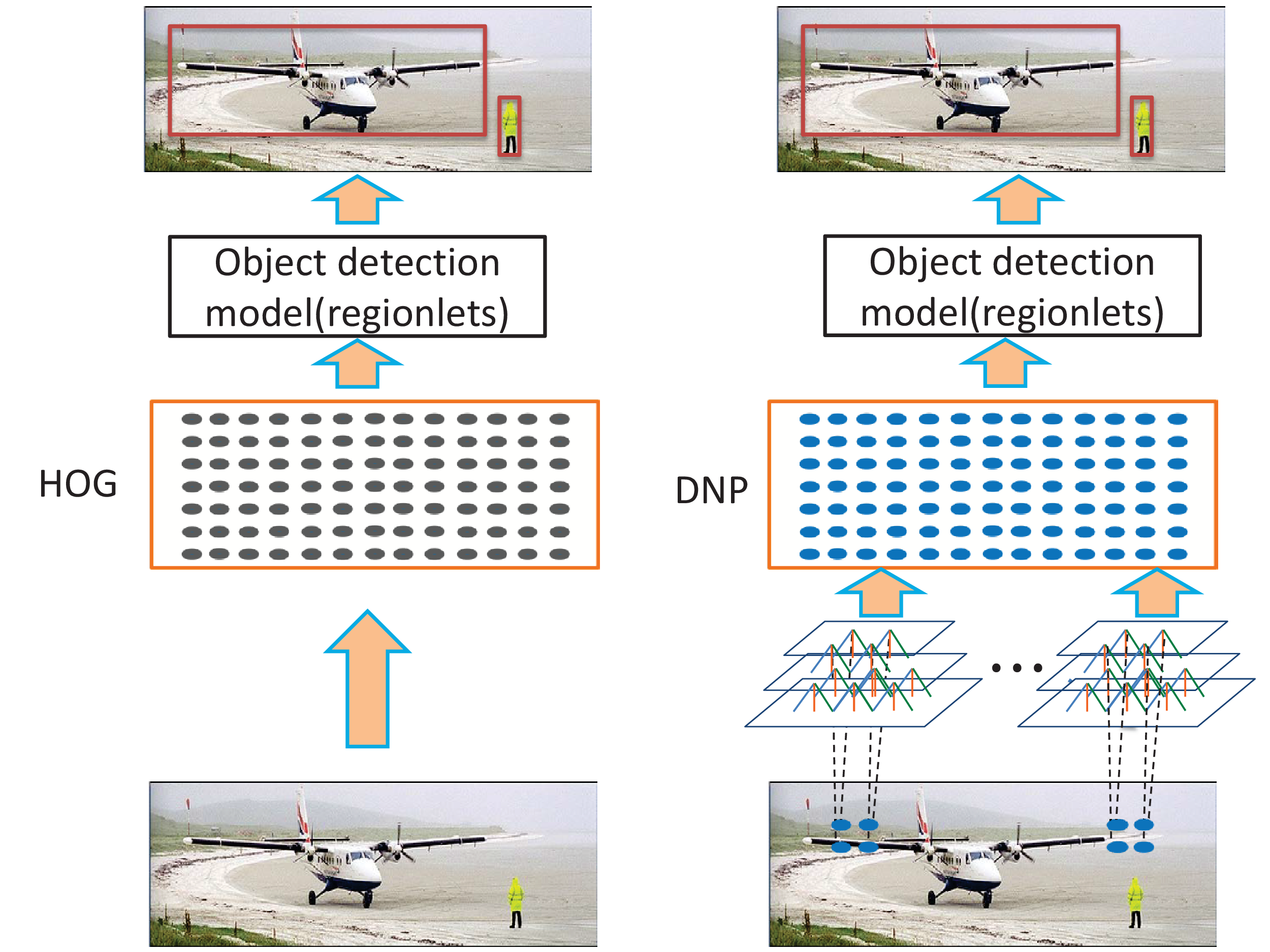} 
\caption{Deep Neural Patterns (DNP) for object detection} 
\label{fig:framework} 
\end{figure} 

These observations motivate us to propose an approach to efficiently incorporate a deep neural
network into conventional object detection frameworks. To that end, we introduce the \emph{Dense
Neural Pattern} (DNP), a local feature densely extracted from an image with arbitrary resolution
using a well trained deep convolutional neural network.  The DNPs not only encode high-level
features learned from a large image data-set, but are also local and flexible like other dense local
features (like HOG or LBP). It is easy to integrate DNPs into the conventional detection frameworks.
More specifically, the receptive field location of a neuron in a deep CNN can be back-tracked to
exact coordinates in the image. This implies that spatial information of neural activations is
preserved. Activations from the same receptive field but different feature maps can be concatenated
to form a feature vector for the receptive field. These feature vectors can be extracted from any
convolutional layers before the fully connected layers. Because spatial locations of receptive
fields are mixed in fully connected layers, neuron activations from fully connected layers do not
encode spatial information. The convolutional layers naturally produce multiple feature vectors that
are evenly distributed in the evaluated image crop ( a $224\times224$ crop for example). To obtain
dense features for the whole image which may be significantly larger than the network input, we
resort to ``network-convolution'' which shifts the crop location and forward-propagate the neural
network until features at all desired locations in the image are extracted.
As the result, a typical PASCAL VOC image only needs to run the neural network several times to
produce DNPs for the whole image depending on the required feature stride, promising low
computational cost for feature extraction.  To adapt our features for the \emph{Regionlets}
framework, we build normalized histograms of DNPs inside each sub-region of arbitrary resolution
within the detection window and add these histograms to the feature pool for the boosting learning
process.  DNPs can also be easily combined with traditional features in the \emph{Regionlets}
framework as explained in Sec.~\ref{sec:regionlet-dnp}.


Our experiments show that the proposed DNPs are very effective and also complementary to traditional
features. On PASCAL 2007 VOC detection benchmark, our framework with \emph{Regionlets} and DNPs achieved
46.1\% mAP compared to 41.7\% with the original \emph{Regionlets}; on PASCAL VOC 2010, it achieves 44.1\%
mAP compared to 39.7\% with the original \emph{Regionlets}. It outperforms the recent approach by
~\cite{Ross2013} with 43.5\% mAP. Furthermore, our DNP features are extracted from the
fifth convolutional layer of the deep CNN without fine-tuning on the target data-set,
while~\cite{Ross2013} used the seventh fully connected layer with fine-tuning. Importantly, for each
PASCAL image, our feature extraction finishes in 2 seconds, compared to approximately 2 minutes from
our replication of~\cite{Ross2013}.

The major contribution of the paper is two-fold: 1) We propose a method to incorporate a
discriminatively-trained deep neural network into a generic object detection framework. This
approach is very effective and efficient. 2) We apply the proposed method to the \emph{Regionlets} object
detection framework and achieved competitive and state-of-the-art performance on the PASCAL VOC
datasets. 

\section{Review of Related Work} 
Generic object detection has been improved over years, due to better deformation modeling, more
effective multi-viewpoints handling and occlusion handling. Complete survey of the object detection
literature is certainly beyond the scope of this paper. Representative works include but not limited
to Histogram of Oriented Gradients~\cite{dalal2005histograms}, Deformable Part-based Model and its
extensions~\cite{lsvm-pami}, \emph{Regionlets}~\cite{wang2013}, \etc.  This paper aims at incorporating
discriminative power of a learned deep CNN into these successful object detection frameworks. The
execution of the idea is based on \emph{Regionlets} object detection framework which is currently the
state-of-the-art detection approach without using a deep neural network. More details about
\emph{Regionlets} are introduced in~\ref{sec:regionlet-dnp}.

More discriminative and robust features are always desirable in object detection, which are arguably
one of the most important domain knowledge developed in computer vision community in past years.
Most of these features are based on colors~\cite{ShahbazCVPR12},
gradients~\cite{dalal2005histograms}, textures~\cite{AphonenPAMI06,wang2009hog}
or relative high order information such as covariance~\cite{Tuzel:2008}. These features are generic
and have been demonstrated to be very effective in object detection. However, none of them encodes
high-level information. The DNPs proposed in this paper complement existing
features in this aspect. Their combination produces much better performance than applying either one
individually. 

Recently, deep learning with CNN has achieved appealing results on image
classification~\cite{Krizhevsky12}.  This impressive result is built on prior work on feature
learning~\cite{lecun1998efficient,hinton2012improving}.  The availability of large datasets like
ImageNet~\cite{Jia2010} and high computational power with GPUs has empowered CNNs to learn deep
discriminative features. A parallel work of deep learning \cite{le2012} without using convolution
also produced very strong results on the ImageNet classification task. In our approach, we choose
the deep CNN architecture due to its unique advantages related to an object detection task as
discussed in Sec.~\ref{sec:network}.  The most related work to ours is~\cite{Ross2013} which
converts the problem of object detection into region-based image classification using a deep
convolutional neural network. Our approach differs in two aspects: 1) We provide a framework to
leverage both the discriminative power of a deep CNN and recently developed effective detection
models. 2) Our method is 74x faster than~\cite{Ross2013}.  There have been earlier work in applying
deep learning to object detection~\cite{ lecun2004learning, le2011haptic}. Among these, most related
to ours is the application of unsupervised multi-stage feature learning for object
detection~\cite{sermanet2012}.  In contrast to their focus on unsupervised pre-training, our work
takes advantage of a large-scale supervised image classification model to improve object detection
frameworks. The deep CNN is trained using image labels on an image classification task.  Learning
deep CNN in an unsupervised manner for our framework may also be interesting but not the current
focus of the paper. 

The proposed approach is a new example of transfer learning, \ie transferring the knowledge learned
from large-scale image classification (in this case, ImageNet image classification) to generic
object detection. There have been some very interesting approaches in transferring the learned
knowledge by deep neural networks. For example, \cite{raina2007self} and \cite{pan2010survey}
illustrated transfer learning with unlabeled data or labels from other tasks. Our work shares a
similar spirit but in a different context. It transfers the knowledge learned from a classification
task to object detection by trickling high-level information in top convolutional layers in a deep
CNN down to low-level image patches. 

\section{Dense Neural Patterns for Object Detection} 

In this section, we first introduce the neural network used to extract dense neural patterns, Then
we provide detailed description of our dense feature extraction approach. Finally, we illustrate the
techniques to integrate DNP with the \emph{Regionlets} object detection framework. 

\subsection{The Deep Convolutional Neural Network for Dense Neural Patterns} \label{sec:network} 
Deep neural networks offer a class of hierarchical models to learn features directly from image
pixels. Among these models, deep convolutional neural networks (CNN) are constructed assuming
locality of spatial dependencies and stationarity of statistics in natural
images~\cite{lecun1995convolutional,Krizhevsky12,ranzato2007sparse}. The architecture of CNNs gives
rise to several unique properties desirable for object detection. Firstly, each neuron in a deep CNN
corresponds to a receptive field~\cite{hubel1968receptive} whose projected location in the image can
be uniquely identified. Thus, the deeper convolutional layers implicitly capture spatial
information, which is essential for modeling object part configurations. Secondly, the feature
extraction in a deep CNN is performed in a homogeneous way for receptive fields at different
locations due to convolutional weight-tying. More specifically, different receptive fields with the
same visual appearance produce the same activations. This is similar to a HOG feature extractor,
which produces the same histograms for image patches with the same appearance. Other architectures
such as local receptive field networks with untied weights (Le et al., 2012) or fully-connected
networks \footnote{Neural networks in which every neurons in the next layer are connected with every
neuron on the previous layer} do not have these properties.  Not only are these properties valid for
a one-layer CNN, they are also valid for a deep CNN with many stacked layers and all dimensions of
its feature maps\footnote{To see this in an intuitive sense, one could apply a ``{\it
network-convolution}'', and abstract the stack of locally connected layers as one layer}.  By virtue
of these desirable properties, we employ the deep CNN architecture.  We build a CNN with five
convolutional layers inter-weaved with max-pooling and contrast normalization layers as illustrated
in Figure 2. In contrast with~\cite{Krizhevsky12}, we did not separate the network into two columns,
and our network has a slightly larger number of parameters. The deep CNN is trained on large-scale
image classification with data from ILSVRC 2010.  To train the neural network, we adopt stochastic
gradient descent with momentum~\cite{lecun1998efficient} as the optimization technique, combined
with early stopping~\cite{girosi1995regularization}. To regularize the model, we found it useful to
apply data augmentation and the dropout technique ~\cite{hinton2012improving,Krizhevsky12}.
Although the neural network we trained has fully connected layers, we extract DNPs only from
convolutional layers since they preserve spatial information from the input image. 

\begin{figure}[tbh] 
\centering 
\includegraphics[width=1.0\columnwidth]{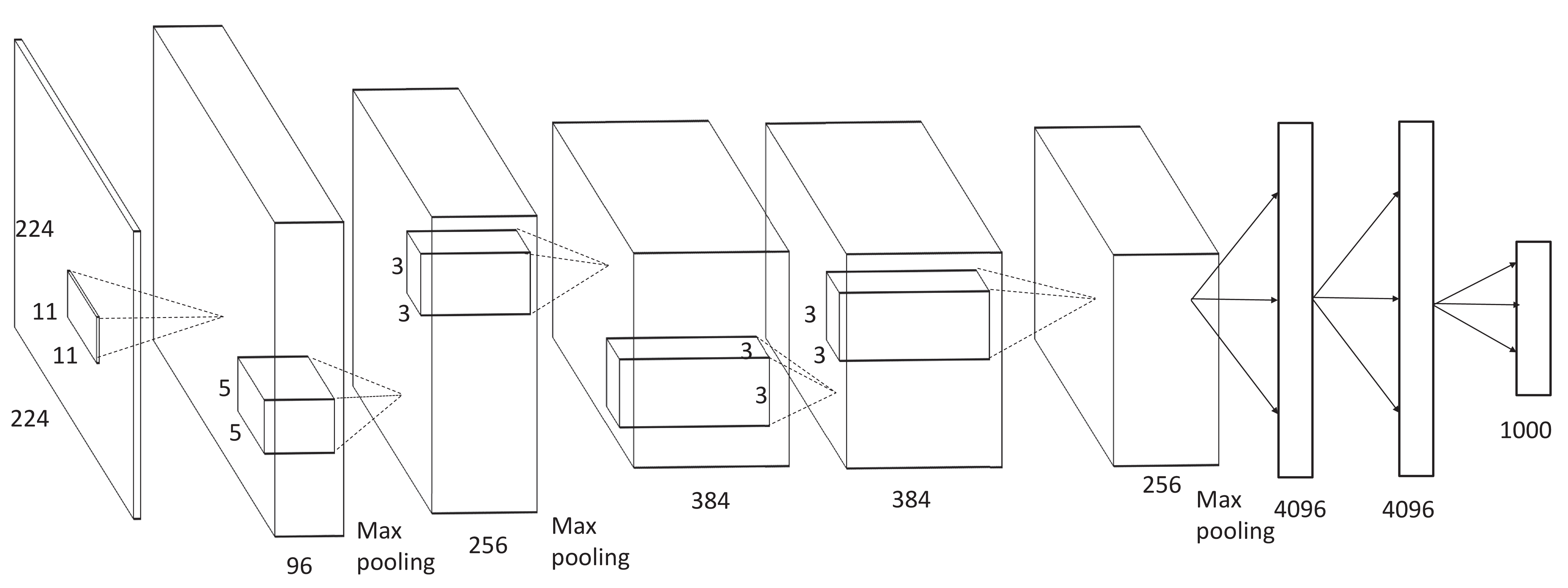} 
\caption{Architecture of the deep convolutional neural network for extracting dense neural
patterns.}
\label{fig:cnn} 
\vspace{-.1in}
\end{figure} 


\subsection{Dense Neural Patterns} \label{sec:DNP_extract} 

After the deep CNN training on large-scale image classification, the recognition module is employed
to produce dense feature maps on high-resolution detection images. We call the combination of this
technique and the resulting feature set Dense Neural Patterns (DNPs). 

\begin{figure}[tbh] 
\centering 
\includegraphics[width=0.9\columnwidth]{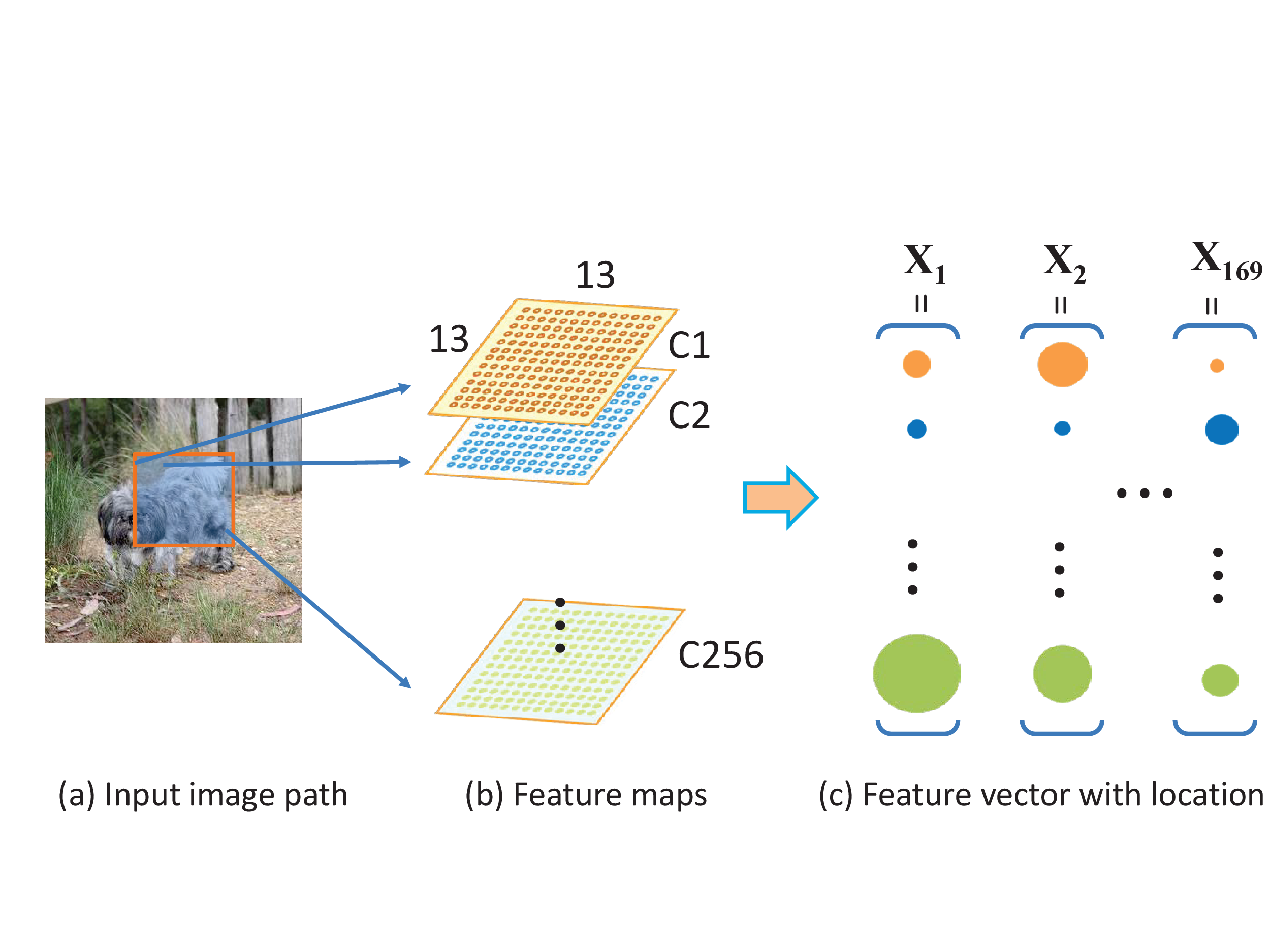} 
\caption{Neural patterns extraction with location association. (a) A square region ($224\times224$)
as the input for the deep neural network. (b) Feature maps generated by filters in the fifth
convolution layer, spatially organized according to their inherited 2-D locations. Each map has
$13\times13$ neural patterns. (c) Feature vector
generated for each feature point. A bigger circle indicates a larger neural activation.}
\label{fig:Layer5} 
\end{figure}

\begin{figure}[tbh] 
\centering 
\includegraphics[width=0.8\columnwidth]{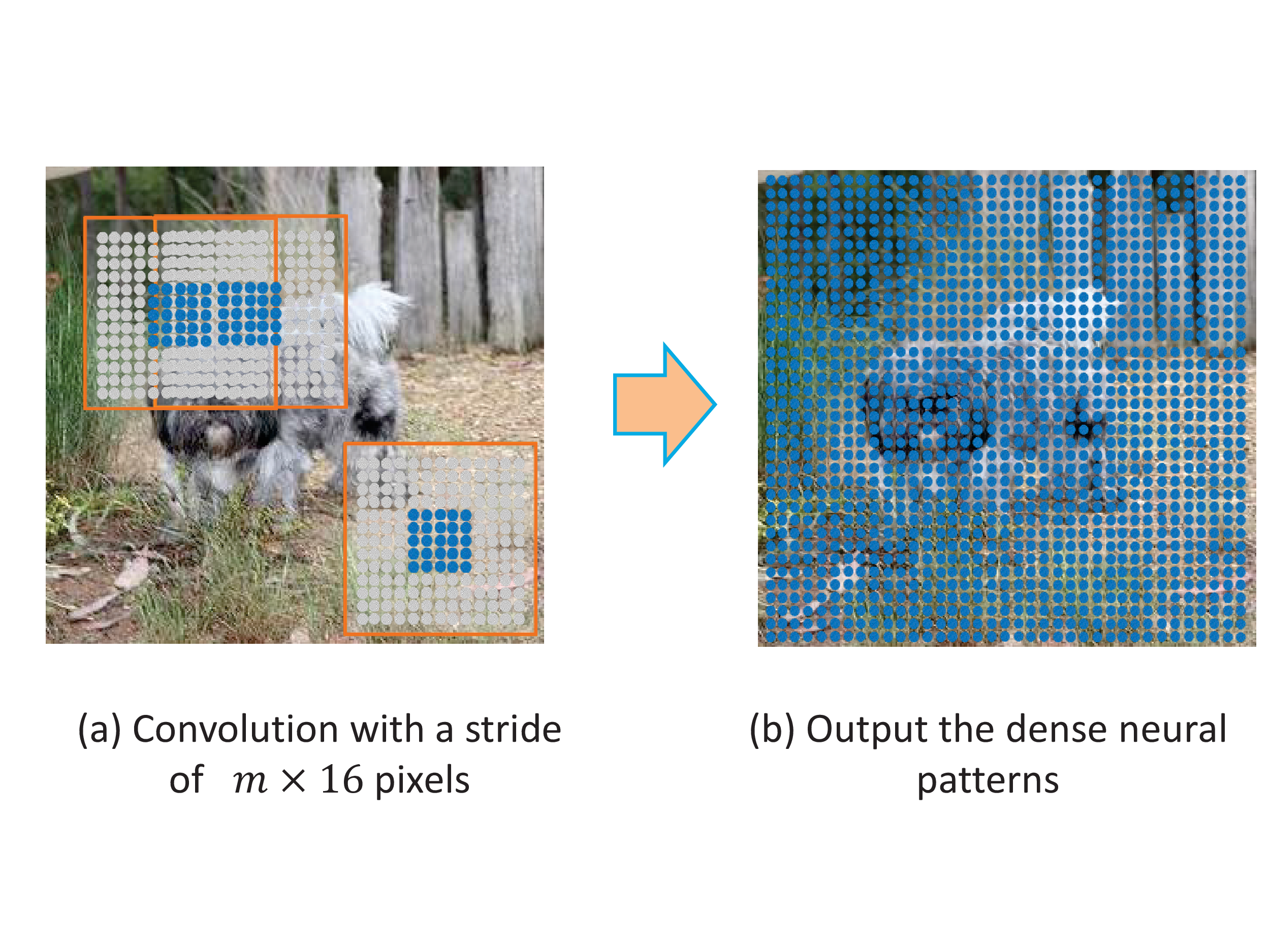} 
\caption{Dense feature maps obtained by shifting the classification window and extract neural patterns
at center positions.} 
\label{fig:densemap} 
\end{figure}

The main idea for extracting dense neural pattern is illustrated in Figure~\ref{fig:Layer5} and
Figure~\ref{fig:densemap}. In the following paragraphs, we first describe the methodologies to
extract features using a deep CNN on a single image patch. Then, we describe the geometries involved
in applying ``{\it network-convolution}'' to generate dense neural patterns for the entire high-resolution
image. 

Each sub-slice of a deep CNN for visual recognition is commonly composed of a convolutional weight
layer, a possible pooling layer, and a possible contrast-normalization layer~\cite{jarrett2009best}.
All three layers could be implemented by convolutional operations. Therefore, seen from the
perspective of preserving the spatial feature locations, the combination of these layers could be
perceived as one convolutional layer with one abstracted kernel. The spatial location of the output
can be traced back by the center point of the convolution kernel. 

As shown in Figure~\ref{fig:Layer5}(b), each convolution kernel produces a sheet of neural patterns.
To tailor dense neural patterns into a flexible feature set for object detectors, we compute the
2-D location of each neural pattern and map it back to coordinates on the original image. As an
example, we show how to compute the location of the top-left neural pattern in
Figure~\ref{fig:Layer5}(b). The horizontal location $x$ of this top-left neural pattern feature is
computed with Equation~\ref{equ:compcenter}:
\begin{equation}
	\label{equ:compcenter}
	x_i = x_{i-1} + (\frac{W_i -1}{2} - P_i)S_{i-1}
\end{equation}
where $i>1$, $x_1=\frac{W_1-1}{2}$, $x_{i-1}$ is the top-left location of the previous layer, $W_i$
is the window size of a convolutional or pooling layer, $P_i$ is the
padding of the current layer, $S_{i-1}$ is the actual pixel stride of two adjacent neural
patterns output by the previous layer which can be computed with Equation~\ref{equ:compstride}
\begin{equation}
	\label{equ:compstride}
	S_i = S_{i-1} \times s_i.
\end{equation}
Here $s_i$ is the current stride using neural patterns output by previous layers as ``pixels''.
Given equation~\ref{equ:compcenter} and equation~\ref{equ:compstride}, the pixel locations of neural
patterns in different layers can be computed recursively going up the hierarchy. Table 1 shows a
range of geometric parameters, including original pixel $x$ coordinates of the top-left neural
pattern and the pixel stride at each layer. Since convolutions are homogeneous in $x$ and $y$
directions, the $y$ coordinates can be computed in a similar manner.  Coordinates of the remaining
neural patterns can be easily computed by adding a multiple of the stride to the coordinates of the
top-left feature point. To obtain a feature vector for a specific spatial location $(x,y)$, we
concatenate neural patterns located at $(x, y)$ from all maps(neurons) as illustrated in
Figure~\ref{fig:Layer5}(c). 

\begin{table}[bht] 
\centering 
\begin{small} 
\caption{Compute the actual location $x_i$ of the top-left neural pattern and the actual pixel
	distance $S_i$ between two adjacent
neural patterns output by layer $i$, based on our deep CNN structure.} 
\label{tab:neurallocation} 

\vspace{1.5mm} 
	\setlength{\tabcolsep}{9.3pt}{
\begin{tabular}{lcccc|cc} 
\toprule
$i$    & Layer & $W_i$ & $s_i$ & $P_i$ & $S_i$ & $x_i$\\
\hline
1      & conv1 & 11    & 4     & 1     & 4     & 6\\
2      & pool1 & 3     & 2     & 0     & 8     & 10\\
3      & conv2 & 5     & 1     & 2     & 8     & 10\\
4      & pool2 & 3     & 2     & 0     & 16     & 18\\
5      & conv3 & 3     & 1     & 1     & 16     & 18\\
6      & conv4 & 3     & 1     & 1     & 16     & 18\\
7      & conv5 & 3     & 1     & 1     & 16     & 18\\
8      & pool3 & 3     & 2     & 0     & 32     & 34\\
\end{tabular} 
}
\end{small} 
\end{table} 

Now that a feature vector can be computed and localized, dense neural patterns can be obtained by
``{\it network-convolution}''. This process is shown in Figure~\ref{fig:densemap}.  Producing dense
neural patterns to a high-resolution image could be trivial by shifting the deep CNN model with
224$\times$224 input over the larger image. However, deeper convolutional networks are usually
geometrically constrained. For instance, they require extra padding to ensure the map sizes and
borders work with strides and pooling of the next layer. Therefore, the activation of a neuron on
the fifth convolutional layer may have been calculated on zero padded values. This creates the
inhomogeneous problem among neural patterns, implying that the same image patch may produce
different activations. Although this might cause tolerable inaccuracies for image classification,
the problem could be detrimental to object detectors, which is evaluated by localization accuracy.
To rectify this concern, we only retain central $5\times5$ feature points of the feature map square.
In this manner, each model convolution generates 25 feature vectors with a $16\times16$ pixel
stride. In order to produce the dense neural patterns map for the whole image using the fifth
convolutional layer, we convolve the deep CNN model every 80 pixels in both $x$ and $y$ direction.
Given a $640\times480$ image, it outputs $40\times 30$ feature points which involves $8\times6$
model convolutions. 

The DNP feature representation has some desirable characteristics which make it substantially
different from and complementary to traditional features used in object detection.

{\bf Robustness to boundary effects caused by local shifts} Most hand-crafted features are not robust
to local shifts due to the hard voting process. Given HOG for example, gradient orientations are
hard voted to spatial($8\times8$) histograms. Features close to the boundary of two feature regions
may be in one region on one example, but the other on another example which causes substantial
feature representation change. The boundary effects may cause difficulties in robust detection.
Moreover, if we shift the window by $8$ pixels, extracted features are completely misaligned.  On
the contrary, the max-pooling in DNPs explicitly handles reasonable pixel shifts. The dense
convolution with shared weights, the data driven learned invariance also implicitly further improve
the robustness to boundary effects and local shifts. 

\begin{figure}[h!] 
\centering 
\includegraphics[width=0.8\columnwidth]{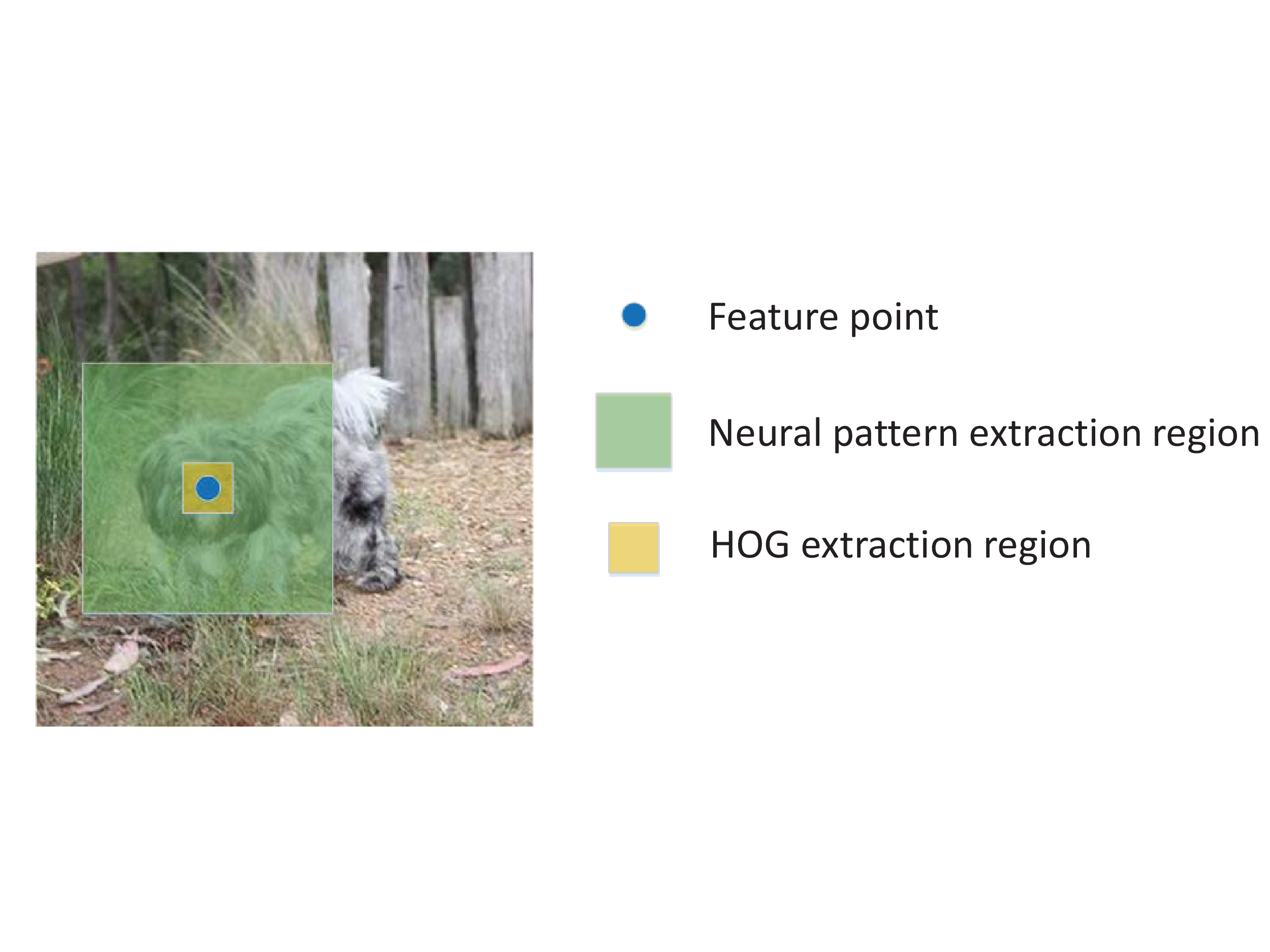} 
\caption{Long-range features for detection from higher layers of
  convolutional networks: The blue circle shows the feature point at which we want to extract
  features. The yellow patch shows the area where HOG features are built (usually $8\times8$).
  The green patch is the receptive field from which the deep net features are are
  extracted ($163\times163$ for the fifth convolutional layer).} 
\label{fig:long-range} 
\vspace{-.1in}
\end{figure} 
{\bf Local features with high-level information} Another significant advantage of DNPs is that the
hierarchical architecture of CNNs offers high-level visual features. More specifically, the features
are indicative of object-level or object-part level visual input. To validate this, we find the
image patches that causes large responses to a selected neural pattern dimension in the deep layers
of the CNN. This visualization is shown in Figure~\ref{fig:det_analysis}. It suggests that patches
which have large feature responses to the same neural pattern dimension correspond to similar
object category, color or contour.  In this respect, DNPs offers significant advantages over
traditional features for object detection.  Details about the visualization can be found in
Sec.~\ref{sec:visualize}.

{\bf Long-range context modeling}
From lower to higher layers, DNP features cover increasingly larger receptive fields. On the fifth
layer, each neuron is responsive to a spatial area of $163\times163$ pixels in the input image. The
features in this layer reacts to appearances of much larger scale as compared to hand-designed local
features like HOG for object detection as shown in Figure~\ref{fig:long-range}. The long-range
effect of the significantly larger context area is beneficial. It is analogous to long-range effects
which were shown to improve localization ~\cite{criminisi2009decision} and image
segmentation~\cite{lezama2011track}.

\subsection{Regionlets with Local Histograms of Dense Neural Patterns}
\label{sec:regionlet-dnp}

\begin{figure*}[tbh]
	\centering

	\includegraphics[width=1.6\columnwidth,height=0.17\textheight]{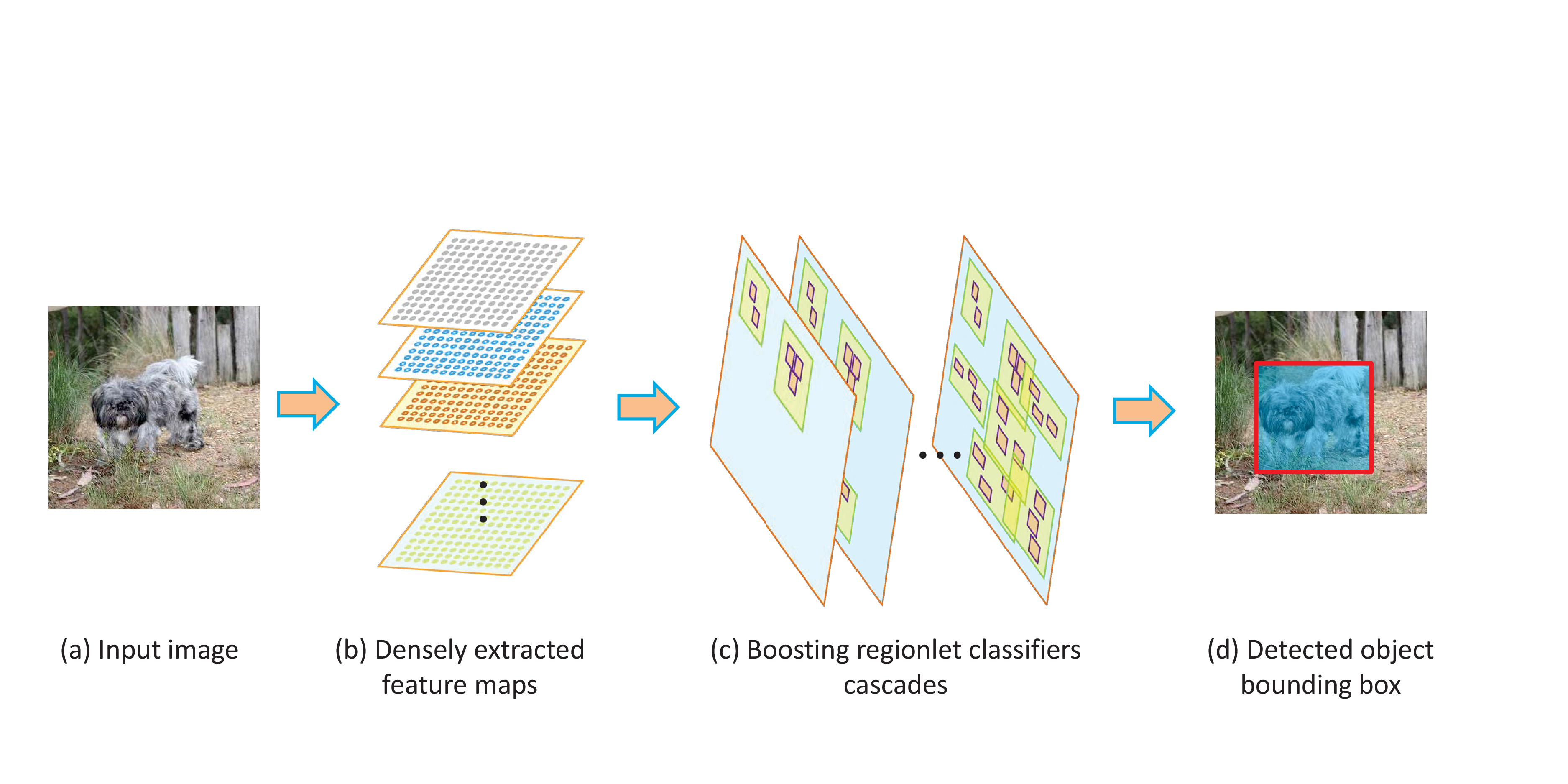}
	\caption{\emph{Regionlets} object detection framework. It learns cascaded boosting classifiers to
		        detect objects of interest. The object searching space is defined using segmentation
				cues. 
			}
			\label{fig:regionlets}
\vspace{-.1in}
\end{figure*}

The \emph{Regionlets} approach for object detection was recently proposed in~\cite{wang2013}.
Compared to classical detection methodologies, which apply a object classifier on dense sliding
windows~\cite{lsvm-pami, dalal2005histograms}, the approach employs candidate bounding boxes from
Selective Search~\cite{van2011segmentation}. Given an image, candidate boxes, \ie, object hypothesis
are proposed using low-level segmentation cues.  

The \emph{Regionlets} approach employs boosting classifier cascades as the window classifier. The
input to each weak classifier is a one-dimensional feature from an arbitrary region $R$. The
flexibility of this framework emerges from max-pooling features from several sub-regions inside the
region $R$.  These sub-regions are named \emph{Regionlets}. In the learning process, the most
discriminative features are selected by boosting from a large feature pool. It naturally learns
deformation handling, one of the challenges in generic object detection. The \emph{Regionlets}
approach offers the powerful flexibility to handle different aspect ratios of objects. The algorithm
is able to evaluate any rectangular bounding box. This is because it removes constraints that come
with fixed grid-based feature extraction. 

\begin{figure}[tbh]
	\centering
	\includegraphics[width=0.8\columnwidth]{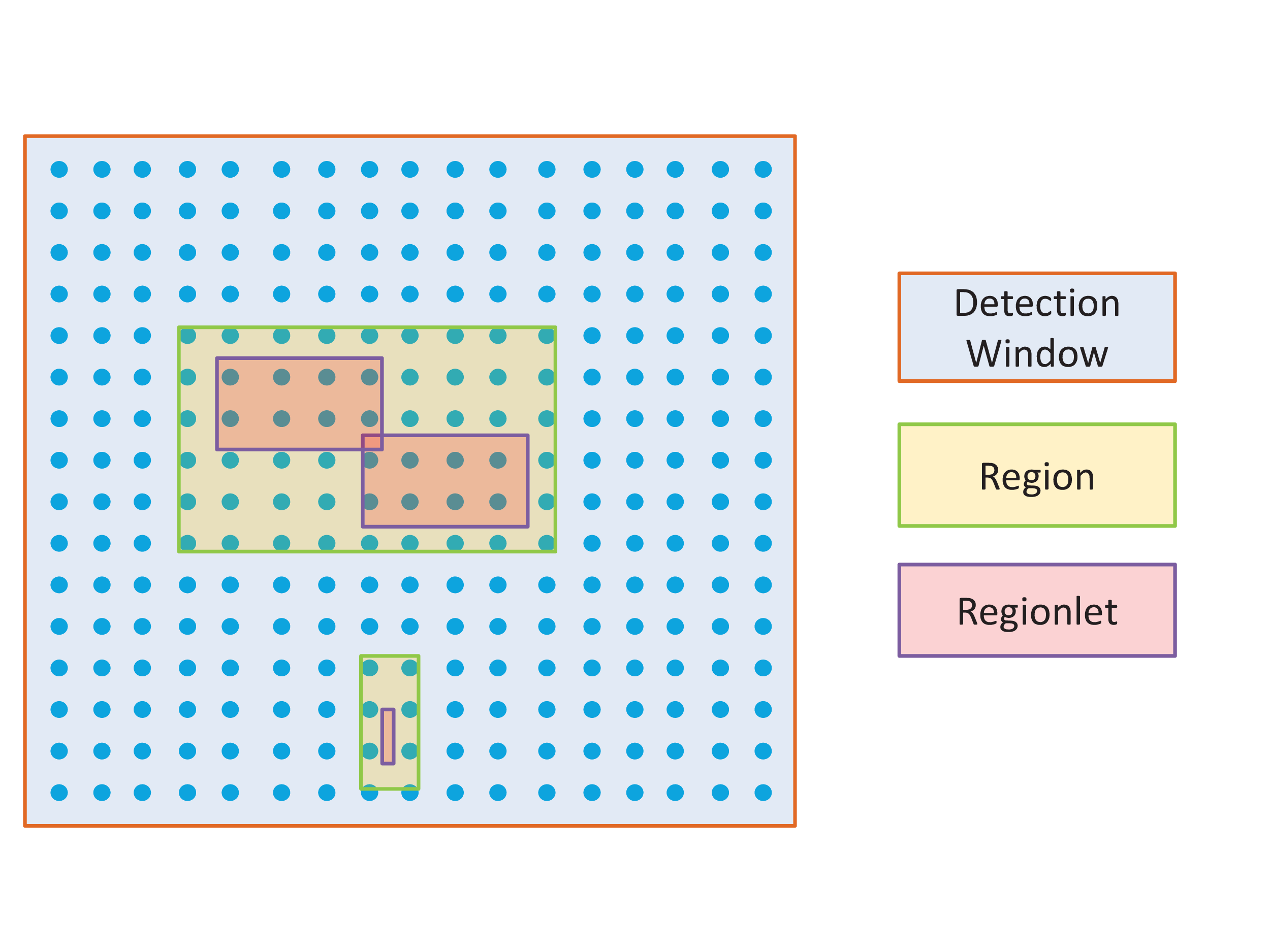}
	\caption{Illustration of feature points, a detection window, regions, and regionlets. Blue
		points represent dense neural patterns extracted in each spatial location. The figure shows that
		a regionlet can spread across multiple feature points, or no feature point.}
	\label{fig:regionlet-feat}
	\vspace{-.1in}
\end{figure}

The dense neural patterns introduced in~\ref{sec:DNP_extract} encode high-level features from a deep
CNN at specific coordinates on the detection image. This makes them a perfect set of features for
the \emph{Regionlets} framework. The basic feature construction unit in the \emph{Regionlets} detection model, \ie
a regionlet, varies in scales and aspect ratios. At the same time, the deep neural patterns from an
image are extracted using a fixed stride which leads to evenly distributed feature points in both
horizontal and vertical directions. As illustrated in Figure~\ref{fig:regionlet-feat}, a regionlet
can cover multiple feature points or no feature point. 
To obtain a fixed length visual representation for a regionlet of arbitrary resolution, we build a
local DNP histogram, or average pooling of DNPs, inside each regionlet. Denote DNPs in a regionlet $r$ as $\{\x_i|i\in(1, \dots
N_r)\}$, where $i$ indicates the index of the feature point, $N_r$ is the total number of feature
points in regionlet $r$. The final feature for $r$ is computed as:
\begin{equation}
	\x = \frac{1}{N_r}\sum_{i=1}^{N_r}\x_i.
\end{equation}
Each dimension of the deep neural patterns corresponds to a histogram bin and their values from
different spatial locations are accumulated inside a regionlet. The histograms are normalized using
L-0 norm.  While most histogram features define a fixed spatial resolution for feature extraction,
our definition allows for a histogram over a region of arbitrary shape and size.
Following~\cite{wang2013}, max-pooling is performed among regionlets to handle local deformations.

To incorporate DNP into the \emph{Regionlets} detector learning framework, in which the weak learner is
based on a 1-D feature, we uniformly sample the {\it DNP}$\times${\it Regionlets} configuration
space to construct the weak classifier pool. Each configuration specifies the spatial configuration
of {\it Regionlets} as well as the feature dimension of {\it DNP}. Because the representation is
1-D, the generated feature pool can be easily augmented to the pool of 
other features such as HOG, LBP or Covariance.

Constructing DNP feature representations for other template-based detectors (similar as HOG
template) is fairly simple. Naturally we just need to concatenate all DNPs in the detection window.
The features can also be directly applied to the Deformable Part-based Model by replacing the HOG
features with the 256 dimensional neural patterns.  

\section{Experiments} 

\begin{table*}[ht] 
\centering 
\begin{small} 
\caption{Detection results on PASCAL VOC 2007 using different layers of neural patterns as the
feature for the \emph{Regionlets} framework.} 
\label{tab:det_results_layers} 
\vspace{1.5mm} 
	\setlength{\tabcolsep}{2.3pt}
	\begin{tabular}{lccccccccccccccccccccc} 
\toprule
        & aero      & bike      & bird      & boat      & bottle & bus       & car  & cat       & chair     & cow       & table     & dog       & horse     & mbike     & person    & plant     & sheep     & sofa      & train     & tv        & \bf{\emph{mAP}} \\
\hline
Layer 1 & 31.5      & 28.9      & 8.1      & 12.7      & 1.2   & 32.9      & 46.3 & 29.6      & 4.2 & 18.3      & 28.2      & 17.2      & 46.3      & 40.6      & 36.1      & 8.2       & 21.6      &17.5      & 41.7      & 25.8      & 24.9 \\
\hline
Layer 2 & 40.1      & 46.2      & 16.6      & 15.6      & 5.3   & 44.3      & 48.7 & 42.8      & 9.9
& 31.6     & 36.2      & 27.8      & 58.1      & 48.4      & 39.8      & 9.1      & 29.6      & 28.7
& 53.6      & 37.2      & 33.5 \\

\hline
Layer 3 & 39.1      & 45.2      & 18.3      & 16.3      & 4.7  & 46.9      & 47.1 & 45.3      & 9.3
& 31.6      & 42.2      & 31.4      & 57.0      & 50.5     & 41.4      & 12.8      & 26.6      &
28.8      & 54.6      & 40.7     & 34.5 \\
\hline
Layer 4 & {\bf 45.0}      & 53.8      & 21.2      & 17.5     & {\bf 8.1}   & {\bf 51.3}      & 50.3 & 52.7      & 12.6
& 32.5      & 44.3      & 39.3      & 62.4      & {\bf 54.8}      & 42.3      & 14.1      & {\bf
33.5}      &
40.8      & 60.3      & 40.9      & 38.9 \\
\hline
Layer 5 & 44.6 & \bf{55.6} & \bf{24.7} & \bf{23.5} & 6.3   & 49.4 & {\bf 51.0} & \bf{57.5} &
\bf{14.3} & \bf{35.9} & \bf{45.9} & \bf{41.3} & \bf{61.9} & 54.7 & \bf{44.1} & \bf{16.0} &
28.6 & \bf{41.7} & \bf{63.2} & \bf{44.2} & \bf{40.2} \\
\hline 
\end{tabular} 
\vspace{-.2in}
\end{small} 
\end{table*} 

To validate our method, we conduct experiments on the PASCAL VOC 2007 and VOC 2010 object detection
benchmarks, following standard evaluation protocols. PASCAL VOC datasets contain 20 categories of
objects. The performance on these datasets is measured by mean average precision (mAP) over all
classes. In the following paragraphs, we describe the experimental set-up, results and analysis for
our object detection approach. 

We train deep neural network with five convolutional layers and three fully connected layers on 1.2
million images in ILSVRC 2010. All input images are center-cropped and resized to $256\times256$ pixels.
The CNN was trained on a NVIDIA Tesla K20c GPU. To improve invariance in our DNP features, we
augment the data with image distortions based on translations and PCA on color channels. After
training for 90 epochs, the deep CNN reached 59\% top 1 accuracy, within a few percent of the
performance in ~\cite{Krizhevsky12} on the ILSVRC 2010 test set. While our aim is to
demonstrate the effectiveness of DNPs in object detection, a deep CNN with better performance is
likely to further improve the detection accuracy. 

The original \emph{Regionlets}~\cite{wang2013} approach utilizes three different features, HOG, LBP and
covariance. In our experiments, we add to the feature pool DNP features from different layers.
During cascade training, 100 million candidate weak classifiers are generated from which we
sample 20K weak classifiers. On each test image, we form proposed object hypothesis
as~\cite{van2011segmentation} and pass them along the cascaded classifiers to obtain final detection
result. 

\subsection{Detection Performance}

We firstly evaluate how the deep neural patterns alone perform with the \emph{Regionlets} framework,
followed with evaluation of the combination of DNP and HOG, LBP, Covariance features. Finally, we
compare our method with other state-of-the-art approaches.

Table~\ref{tab:det_results_layers} presents the detection performance using dense neural patterns
extracted from different layers of the deep convolutional neural network on PASCAL VOC 2007 dataset.
It shows that the performance increases with respect to the layer hierarchy. DNPs from the fourth layer
and the fifth layer have similar performance, both of which are much better then those from lower
layers.

Table~\ref{tab:detres_dnphog} summarizes the performance(sorted in ascending order) of traditional
features, DNP and their combinations on PASCAL VOC 2007. It is interesting that DNPs from the second
layer and third layer have comparable performance with the well engineered features such as HOG, LBP
and Covariance features. DNPs from the fifth layer outperforms any single features, and are
comparable to the combination of all the other three features. The most exciting fact is that DNPs
and hand-designed features are highly complementary. Their combination boosts the mean average
precision to 46.1\%, outperforming the original Reginolets approach by 4.4\%.  Note that we did not
apply any fine-tuning of the neural network on the PASCAL dataset. 

The combination of DNPs and hand-crafted low-level features significantly improves the detection
performance.  As aforementioned, low-level DNPs perform similarly as HOG.  To determine whether the
same synergy can be obtained by combining low-level and high-level DNPs, we combine the DNPs from
the fifth convolutional layer and the second convolutional layer. The performance is shown in
Table~\ref{tab:detres_combine}.  However, the combination only performs slightly better (0.2\%) than
using the fifth layer only. This may be because the fifth layer features are learned from the lower
level which makes these two layer features less complementary.

\begin{table}
\centering 
\begin{small} 
\caption{Detection results using traditional feature and Deep Neural Patterns on PASCAL VOC 2007.
The combination of traditional features and DNP shows significant improvement.} 
\label{tab:detres_dnphog} 

\vspace{1.5mm} 
\begin{tabular}{lc} 
\toprule
Features & Mean AP\\ 
\hline 
DNP Layer 1 & 24.9 \\ 
\hline 
DNP Layer 2 & 33.5 \\ 
\hline 
LBP & 33.5 \\ 
\hline 
Covariance & 33.7 \\ 
\hline 
DNP Layer 3 & 34.5 \\ 
\hline 
HOG & 35.1 \\ 
\hline 
DNP Layer 4 & 38.9 \\ 
\hline 
DNP Layer 5 & 40.2 \\ 
\hline 
HOG, LBP, Covariance & 41.7 \\ 
\hline 
{\bf HOG, LBP, Covariance, DNP Layer 5} & {\bf 46.1}\\
\hline 
\end{tabular} 
\end{small} 
\end{table} 

\begin{table}
\centering 
\begin{small} 
\caption{Performance comparison between two feature combination strategies: 1) Combination of neural
patterns from the fifth layer and neural patterns from a shallow layer(second layer). 2) Combination of neural patterns
from the fifth layer and hand-crafted low-level features. } 
\label{tab:detres_combine} 

\vspace{1.5mm} 
\begin{tabular}{lc} 
\toprule
Features & Mean AP\\ 
\hline 
DNP Layer 5 & 40.2\% \\ 
\hline 
DNP Layer 5 + Layer 2& 40.4\% \\ 
\hline 
DNP Layer 5 + HOG, LBP, Covariance& 46.1\% \\ 
\hline 
\end{tabular} 
\vspace{-.1in}
\end{small} 
\end{table}


\begin{table}[bht] 
\centering 
\begin{small} 
	\caption{Detection results(mean average precision\%) on PASCAL VOC 2007 and VOC 2010 datasets. {\bf DPM:} Deformable Part-based
	Model~\cite{lsvm-pami};{\bf SS\_SPM:} Selective Search with Spatial Pyramid
	Matching~\cite{van2011segmentation};{\bf Objectness:}~\cite{alexe2012measuring};
	{\bf BOW:}~\cite{vedaldi2009multiple}; {\bf Regionlets:}Regionlets method with HOG, LBP
	Covariance feature~\cite{wang2013}, {\bf DNP+ Regionlets:}Regionlets method with HOG, LBP
	Covariance feature and DNPs.{\bf R-CNN pool$_5$}: Region based classification for
	detection~\cite{Ross2013} with features from the fifth convolutional layer with max pooling.
	{\bf {R-CNN FT fc$_7$:} } Region based classification for
	detection~\cite{Ross2013} with features from the fully connected layer fine-tuned on the PASCAL
VOC datasets.}
\label{tab:det_results_07} 
\vspace{1.5mm} 
	\setlength{\tabcolsep}{9.5pt}
\begin{tabular}{l|c|c} 
	\toprule
& VOC 2007 & VOC2010 \\ 
\hline 
DPM & 33.7 & 29.6\\
\hline 
SS\_SPM & 33.8 & 34.1\\
\hline 
Objectness & 27.4 & N/A\\
\hline 
BOW & 32.1 & N/A\\
\hline 
Regionlets & 41.7 & 39.7\\
\hline 
R-CNN pool$_5$& 40.1 & N/A\\
\hline 
R-CNN FT fc$_7$ & {\bf 48.0} & 43.5 \\
\hline 
DNP+Regionlets & 46.1 & {\bf 44.1}\\
\hline 
\end{tabular} 
\end{small} 
\end{table} 

Table~\ref{tab:det_results_07} shows detection performance comparison with other detection methods
on PASCAL VOC 2007 and VOC 2010 datasets. We achieved 46.1\% and 44.1\% mean average precision on
these two datasets which are comparable with or better than the current stat of the art
by~\cite{Ross2013}.  Here we compare to results with two different settings in~\cite{Ross2013}:
features from the fifth convolutional layer after pooling, features from the seventh fully connected
layer with fine-tuning on the PASCAL datasets. The first setting is similar to us except that
features are pooled. Our results are better(46.1\% vs 40.1\% on VOC 2007) than~\cite{Ross2013} on
both datasets in this setting. The approach in~\cite{Ross2013} requires resizing a candidate region
and apply the deep CNN thousands of times to extract features from all candidate regions in an
image.  The complexity of our method is independent of the number of candidate regions which makes
it orders of magnitude faster.  Table ~\ref{tab:comp_speed} shows the comparison
with~\cite{Ross2013} in terms of speed using the first setting.\footnote{The time cost of the second
	setting in~\cite{Ross2013} is higher because of the computation in fully connected layer.} The
	experiment is performed by calculating the average time across processing all images in the
	PASCAL VOC 2007 dataset.  DNPs extraction take 1.64 seconds per image while~\cite{Ross2013}
	requires 2 minutes. The numbers are obtained on an Intel Xeon CPU E5-2450 blade server.

\begin{table}[bht] 
\centering 
\begin{small} 
\caption{Speed comparison with directly extracting CNN features for object candidates~\cite{Ross2013} . } 
\label{tab:comp_speed} 

\vspace{1.5mm} 
\begin{tabular}{ccc} 
\toprule
& {\bf R-CNN pool$_5$}& {\bf Ours}\\ 
\hline 
Resize object candidate regions & Yes & No\\ 
\hline 
Number of model convolutions & $\sim$ 2213 & $\sim$ 30\\ 
\hline 
Feature extraction time per image& 121.49s & {\bf 1.64}s\\ 
\hline 
\end{tabular} 
\end{small} 
\end{table}

\subsection{Visual Analysis } 
\label{sec:visualize}

Is the increase in detection performance by adding dense neural patterns attributable to the
high-level cues encoded by DNPs? To answer this question, we devise a visualization techniques for
the most important features used by the detector. The learning process for boosting selects
discriminative weak classifiers. The importance of a feature dimension roughly corresponds to how
frequently it is selected during training. We count the occurrence of each dimension in the final
weak classifier set and find the DNP feature dimension most frequently selected by boosting. To
visualize these feature dimensions, we retrieve image crops from the dataset which give the highest
responses to the corresponding neurons in the deep CNN.

\begin{figure}[h!] 
\centering 
\includegraphics[width=1.0\columnwidth]{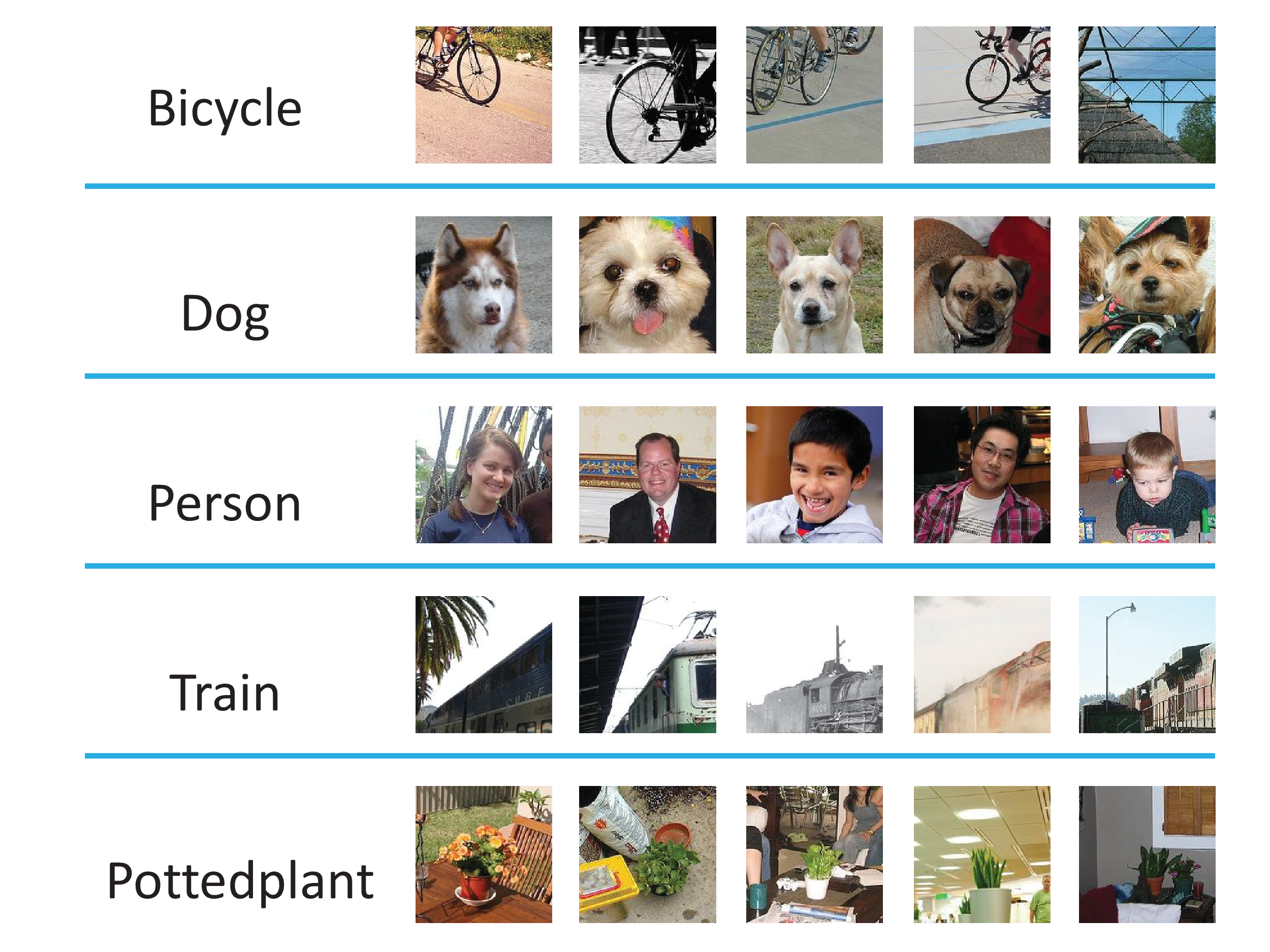} 

\caption{Visualization of the high-level information encoded by neural patterns from the fifth
	convolutional layer. The patches are obtained by: 1) Determine the most frequently selected
	neural pattern dimension (1 out of 256) for an object category. 2) Run the neural pattern extractor as a
	detector, using the value of the extracted neural patterns as detection scores. 3) Collect and
	rank detection results, visual patches with larger neural pattern values are ranked top.}

\label{fig:det_analysis} 
\end{figure}

Figure ~\ref{fig:det_analysis} shows the visualization. The ideal case is that the most frequent
neural patterns selected in a person detector give high responses to parts belonging to a person.
This indicates that the neural patterns encode high-level information. The left column of
Figure~\ref{fig:det_analysis} describes the object
category we want to detect. Right columns show visual patches which give high responses to the most
frequently selected neural pattern dimension for the category. This analysis indicates that the
selected neural patterns encode part-level or object-level visual features highly correlated with
the object category. For a dog detector, neural patterns related to a dog face are frequently
selected. We also performed a similar analysis with the HOG feature. In comparison, the frequently
selected HOG dimension carries a lot less categorical information because gradients are low-level
visual features.

\section{Conclusion} 

In this paper, we present a novel framework to incorporate a discriminatively trained deep
convolutional neural network into generic object detection. It is a fast effective way to enhance
existing conventional detection approaches with the power of a deep CNN. Instantiated with
\emph{Regionlets} detection framework, we demonstrated the effectiveness of the proposed approach on
public benchmarks. We achieved comparable performance to state-of-the-art with 74 times faster speed
on PASCAL VOC datasets. We also show that the DNPs are complementary to traditional features used in
object detection. Their combination significantly boosts the performance of each individual feature.

\subsection*{Acknowledgments}

This work was done during the internship of the first author at NEC Laboratories America in
Cupertino, CA. 

{\footnotesize
\bibliographystyle{ieee} 
\bibliography{deeplearning}
}
\end{document}